\documentclass{article}

\PassOptionsToPackage{numbers}{natbib}

\usepackage[preprint]{neurips_2020}

\usepackage[utf8]{inputenc} 
\usepackage[T1]{fontenc}    
\usepackage{hyperref}       
\usepackage{url}            
\usepackage{booktabs}       
\usepackage{amsfonts}       
\usepackage{nicefrac}       
\usepackage{microtype}      
\usepackage[usenames,dvipsnames]{xcolor}
\usepackage{bbm}
\usepackage{amsmath}
\usepackage{amssymb}
\usepackage{amsthm}
\usepackage{subcaption}
\usepackage{tabularx}

\usepackage{todonotes} 

\newboolean{include-notes}
\setboolean{include-notes}{true}

\newcommand{\TODO}[1]{\ifthenelse{\boolean{include-notes}}
 {{\color{red} TODO: #1}}{}}
\newcommand{\Rohin}[1]{\ifthenelse{\boolean{include-notes}}
 {{\color{ForestGreen} RS: #1}}{}}
\newcommand{\Micah}[1]{\ifthenelse{\boolean{include-notes}}
 {{\color{orange} MC: #1}}{}}
\newcommand{\adnote}[1]{\ifthenelse{\boolean{include-notes}}
 {{\color{blue}AD: #1}}{}}
 \newcommand{\Paul}[1]{\ifthenelse{\boolean{include-notes}}
 {{\color{purple} #1}}{}}

\newcommand{\dist}[1]{\Delta(#1)}
\newcommand{\tuple}[1]{\langle #1 \rangle}
\newcommand{\set}[1]{\{ #1 \}}
\newcommand{\expect}[2]{\mathbb{E}_{#1}\left[{#2}\right]}

\newcommand{\M}{\mathcal{M}}
\newcommand{\St}{\mathcal{S}}
\newcommand{\A}{\mathcal{A}}
\newcommand{\T}{\mathcal{T}}
\newcommand{\R}{R} 
\newcommand{\PS}{\mathcal{P}}
\newcommand{\Reals}{\mathbb{R}}

\newcommand{\prg}[1]{\noindent\textbf{#1}}

\title{Evaluating the Robustness of Collaborative Agents}

\author{%
  Paul Knott\thanks{Alternative email: knottquantum@gmail.com} \\
  University of Nottingham \\
  \texttt{paul.knott@nottingham.ac.uk} \\
  
   \And
   Micah Carroll \\
   UC Berkeley \\
   \texttt{mdc@berkeley.edu} \\
   \AND
   Sam Devlin \\
   Microsoft Research \\
   \And
   Kamil Ciosek \\
   Microsoft Research \\
   \And
   Katja Hofmann \\
   Microsoft Research \\
   \And
   A. D. Dragan \\
   UC Berkeley \\
   \And
   Rohin Shah \\
   UC Berkeley \\
}

\begin{document}

\maketitle

\begin{abstract}
In order for agents trained by deep reinforcement learning to work alongside humans in realistic settings, we will need to ensure that the agents are \emph{robust}. Since the real world is very diverse, and human behavior often changes in response to agent deployment, the agent will likely encounter novel situations that have never been seen during training. This results in an evaluation challenge: if we cannot rely on the average training or validation reward as a metric, then how can we effectively evaluate robustness? We take inspiration from the practice of \emph{unit testing} in software engineering. Specifically, we suggest that when designing AI agents that collaborate with humans, designers should search for potential edge cases in \emph{possible partner behavior} and \emph{possible states encountered}, and write tests which check that the behavior of the agent in these edge cases is reasonable. We apply this methodology to build a suite of unit tests for the Overcooked-AI environment, and use this test suite to evaluate three proposals for improving robustness. We find that the test suite provides significant insight into the effects of these proposals that were generally not revealed by looking solely at the average validation reward.

\end{abstract}

\section{Introduction}

Deep reinforcement learning (deep RL) has been used very successfully to train agents that perform very well in the average case~\cite{berner2019dota, vinyals2019alphastar, silver2017mastering}. However, deployment of an agent in the real world will often have stringent robustness requirements~\citep{lohn2020estimating, gasparik2018safety}. Due to the diversity of the real world, a deployed agent will encounter many situations and human behaviors that were never seen during development. Recent work has shown failures of policies learned from deep RL in simulation~\cite{gleave2019adversarial}, suggesting that we do not get such robustness by default. 

We are particularly interested in building agents that \emph{collaborate} with humans in order to help them accomplish their goals, a setting that has recently been tackled with deep RL~\cite{hu2020other,carroll2019utility,lerer2019learning}. There are several approaches we could use to improve the robustness of such agents. For example, rather than training a deep RL agent to play with a single human model trained with behavior cloning~\citep{carroll2019utility}, we can potentially improve 1) the \emph{quality} of the human model by incorporating Theory of Mind (ToM) \citep{choudhury2019utility}, 2) the \emph{human model diversity} by using a population of human models, and 3) the \emph{state diversity} by, e.g., initializing from states visited in human-human gameplay.

However, it is hard to \emph{evaluate} these ideas. Since we care about robustness to novel situations, the average reward on the training distribution is not a sufficient metric. We would like to know the true distribution over performance during deployment, but this is never available because the deployment of the agent itself changes the distribution of inputs it receives~\citep{Scholz}. Even pairing the learned agent with people in a user study is not usually representative of the performance at deployment time, since in many realistic domains there is a long tail of unusual edge cases that would likely not be seen for any reasonable sample size. However, at deployment time, if the user base is large enough, the likelihood of eventually running into one of these edge cases is greatly increased, and could have large consequences depending on the extent of the failure and the deployment setting.

Software engineering faces a similar challenge, in which incorrect programs often give the right output on the vast majority of possible inputs, but fail on a few specific edge cases. The most widespread technique to combat this issue is \emph{testing}, in which the programmer explicitly writes down potential edge cases and their expected outputs. While unit testing does not guarantee correctness, even automatically generated unit tests can find many issues~\citep{shamshiri2015do}.

\begin{figure*}[t]
    \centering
    \begin{subfigure}[t]{0.32\textwidth}
        \centering
        \includegraphics[height=0.68\linewidth]{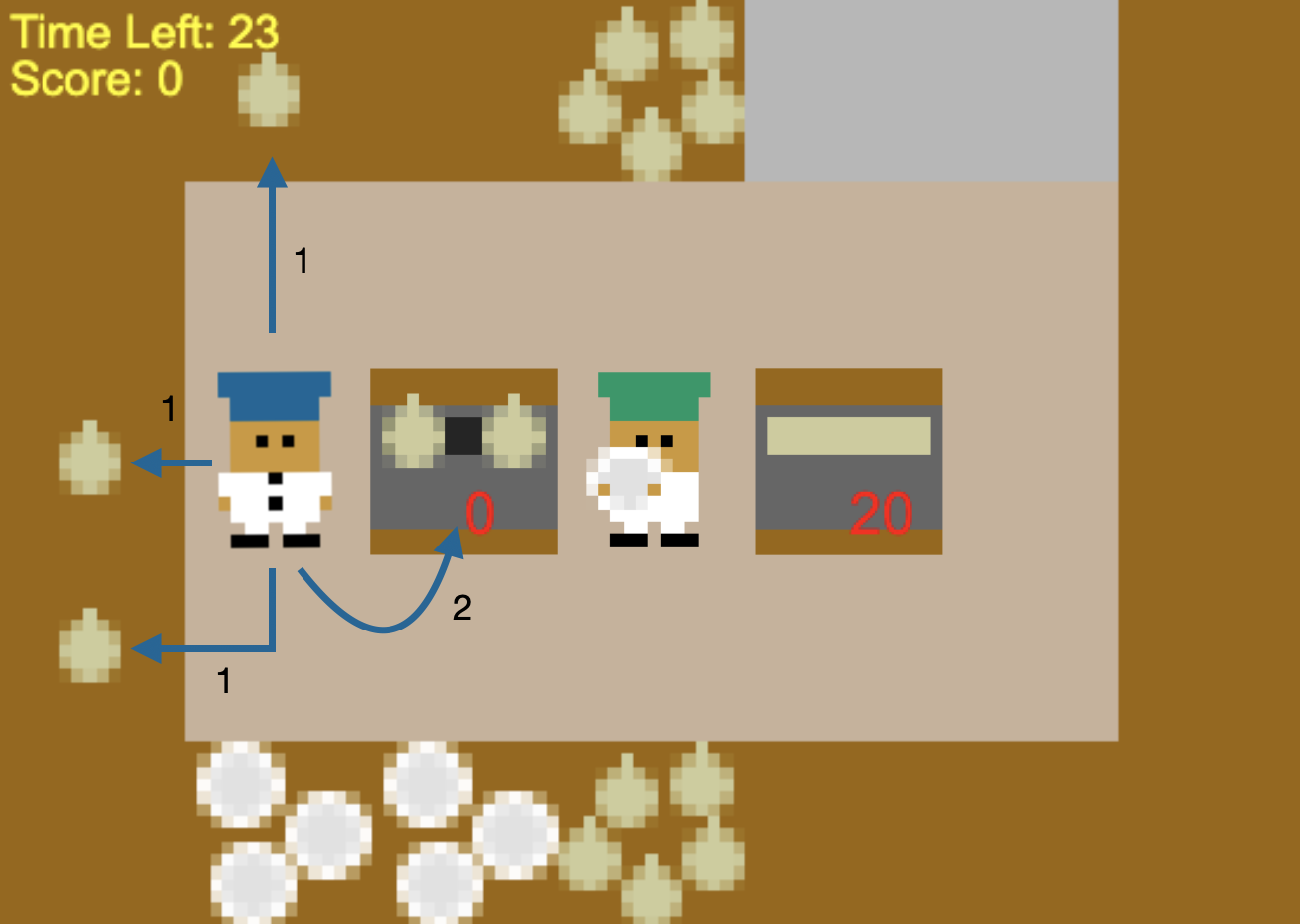}
        \caption{Example \emph{state robustness} unit test. Since its partner (green) is already holding a plate, the agent (blue) is in the best position to get an onion for the left pot, regardless of how its partner plays.}
        \label{fig:state-robustness}
    \end{subfigure}
    \hfill
    \begin{subfigure}[t]{0.32\textwidth}
        \centering
        \includegraphics[height=0.68\linewidth]{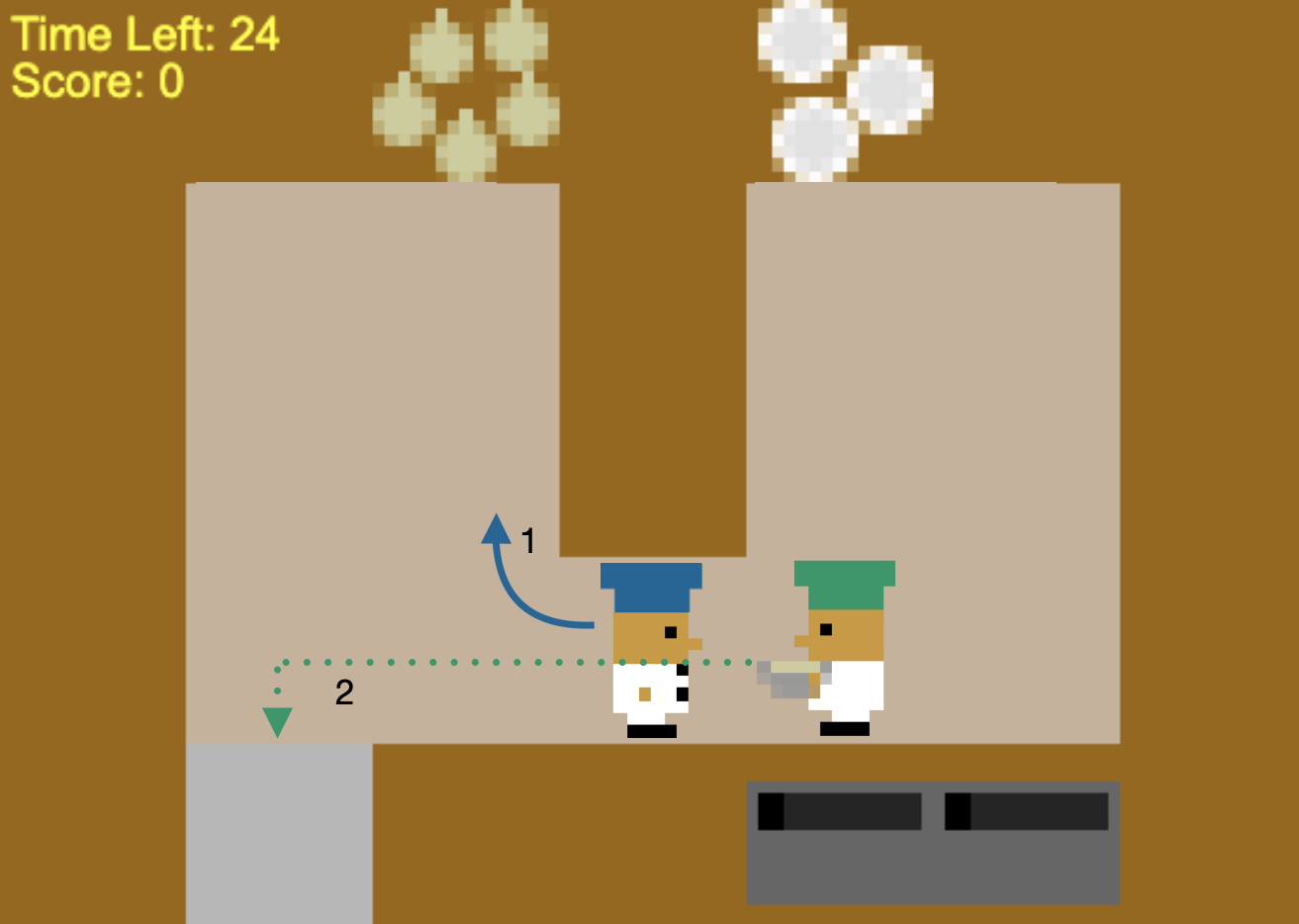}
        \caption{Example \emph{agent robustness} unit test. The agent's partner (green) stubbornly insists on delivering the soup in its hands, and so the agent (blue) should get out of the way.}
        \label{fig:agent-robustness}
    \end{subfigure}
    \hfill
    \begin{subfigure}[t]{0.32\textwidth}
        \centering
        \includegraphics[height=0.68\linewidth]{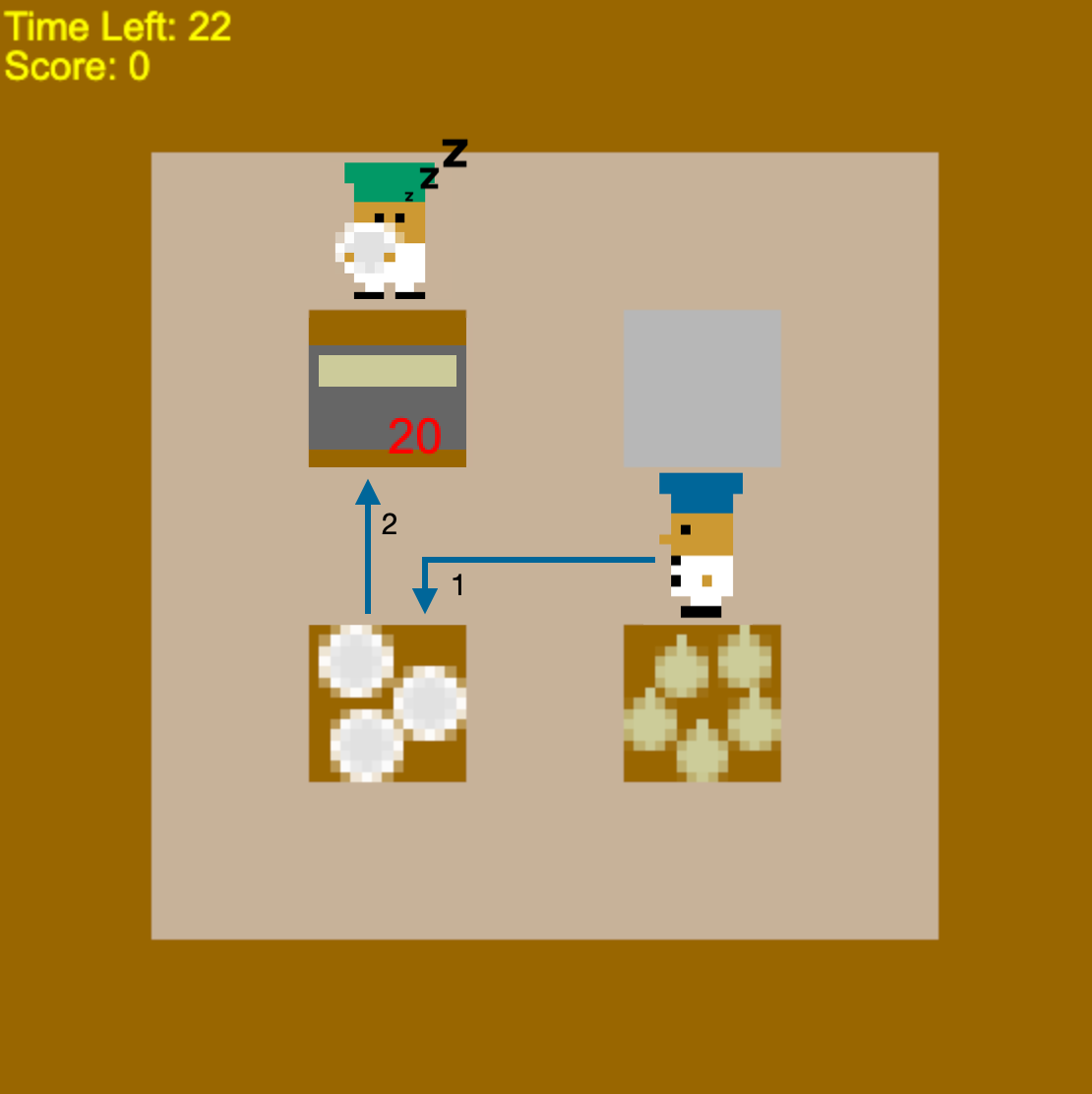}
        \caption{Example \emph{agent \& memory robustness} unit test. The fact that the agent's partner (green) is AFK, and thus that the agent (blue) should pick up and deliver the soup themselves, is only possible to detect through memory.}
        \label{fig:agent-memory-robustness}
    \end{subfigure}
    \caption{Example unit tests for evaluating robustness of agents trained to play Overcooked. Note that in these cases (and all other unit tests), the correct behavior for the evaluated agent is unambiguous – any truly robust agent should be able to successfully complete the task the test is checking for.}
    \label{fig:unit-tests}
\end{figure*}


In this paper, we develop a methodology for applying the testing paradigm to human-AI collaboration, and we demonstrate its utility on the two-player Overcooked environment~\cite{carroll2019utility,Overcooked}, in which players control chefs in a kitchen to cook and serve dishes. A good test suite should test potential edge cases in the \emph{states} that the agent should be able to handle (e.g. what if a couple of plates were accidentally left on the kitchen's counters?), and the \emph{partners} that the agent should be able to play with (e.g. what if the partner were to stay put for a while because they are thinking or away from their keyboard?). Additional examples in Overcooked are shown in Figure~\ref{fig:unit-tests}.

The benefit of the test suite is that it can give us more information than we could get by observing the reward alone. We demonstrate such benefits within Overcooked. First, we confirm the canonical wisdom that agents trained via vanilla deep RL are not robust. We then evaluate the three proposals above on improving state diversity, human model diversity, and human model quality. Our results show that the test suite provides significant insight into robustness that is not very correlated with the information provided by the average validation reward: for example, we find that improving state diversity does improve robustness as measured by our test suite, but this comes at the cost of a \emph{decrease} in average validation reward.

A given test suite will never be final: there will likely always be more edge cases to include. As different types of failure modes are found or imagined, they can be added into the test suite. We do not claim that unit testing allows us to achieve perfect robustness: rather, we see them as a major improvement over the current status quo of evaluating reward on random rollouts (which only tests the edge cases that are encountered randomly). Current deep RL agents are clearly not robust -- none of the agents we tested scored above 65\% in Overcooked -- suggesting that our approach can serve as a useful metric for the foreseeable future. Once agents routinely get high scores on the tests, we should consider how to use additional human effort to create even better robustness metrics.

\section{Related work}

\prg{Evaluation strategies.} Many disciplines must contend with the challenges of real-world deployment. The field of \emph{safety engineering} in particular develops best practices for building systems that can operate safely and robustly in the real world, and pays particular attention to the role of human factors~\citep{leveson2016engineering}. Our process of building test suites can be thought of as an exercise in explicitly mapping out possible variation in human factors in order to increase our confidence in the agents' ability to respond to this variation.

Within human-AI collaboration, one evaluation strategy is to search \emph{adversarially} for situations in which the agent fails~\citep{uesato2018rigorous}. However, in collaborative environments it is unclear how to design such an adversary. Giving the adversary arbitrary control over the partner behavior is far too strong a requirement: for example, in Overcooked the adversary could simply stand at the soup delivery location, preventing soups from getting delivered and guaranteeing that no reward is achieved.

Our work continues an increasing trend in machine learning in which simple, easily calculated metrics are insufficient to capture performance, and we must instead evaluate results based on human judgment~\citep{stiennon2020learning,adiwardana2020towards,zhou2019hype}. 

\prg{Human-AI collaboration.} We focus on building agents that can collaborate with humans by pairing the agents with different human models during training. Recent work has proposed two other options. \emph{Other play}~\cite{hu2020other} makes the assumption that the agent should be invariant to permutations of symmetries of the game, in order to enable zero-shot collaboration (i.e. collaboration without any human data). However, this will not necessarily result in optimal behavior given specific collaborators. Alternatively, human data can be used to discover which of several Nash equilibria humans play, and bias an agent trained in self-play to learn the same equilibrium~\cite{lerer2019learning,tucker2020adversarially}. However, these techniques must make strong assumptions about human gameplay, which may not hold in more complex settings. 

\prg{Training robust agents.} Much recent work has focused on creating agents that are robust to variation in the environment, especially for sim-to-real transfer. One common technique is \emph{domain randomization}, where some aspect of the environment is varied randomly in order to learn a policy that can robustly succeed regardless of that aspect~\cite{openai2019rubiks, peng2018sim, yu2017preparing, tan2018sim}. Our populations can be thought of as a domain randomization technique, where the randomization is done over the parameters of the ToM or the initialization of the BC model parameters. Populations of agents have in fact been used to improve average case reward, especially in zero-sum settings~\cite{vinyals2019alphastar,jaderberg2019human}. Typically, the agents in the population are all trained in self-play with each other. While this is effective in competitive environments, it tends to converge to overly specific Nash equilibria in collaborative environments~\cite{carroll2019utility}. Another common approach to achieve robustness is to train the agent with a constrained \emph{adversary} that attempts to sabotage the agent~\cite{pinto2017robust,chalaki2019zero, pattanaik2018robust, shen2019robust, uesato2018rigorous}. Unfortunately, it is unclear how to apply such a technique for human-AI collaboration: an unconstrained adversarial human model would often be able to prevent any reward from being accumulated, while a constrained model may not generalize to real humans. 

\prg{Overcooked-AI.} Recent work~\cite{wang2020too, carroll2019utility} has used environments based on the popular video game \emph{Overcooked}~\cite{Overcooked}, in which players control chefs in a kitchen to cook and serve dishes. We use the Overcooked-AI environment implementation from \citet{carroll2019utility}, in which the only objects are onions, dishes, and soups. Players work together to place 3 onions in a pot, leave them to cook for 20 timesteps, put the resulting soup in a dish, and serve it. All players are given reward (20 points) exclusively when a soup is served.

\section{Preliminaries}

We introduce the formal setting. Note that we denote the space of distributions over $X$ as $\dist{X}$.

\subsection{Multiagent MDPs} \label{sec:multiagent-mdp}

An \emph{$n$-player Markov Decision Process} $\M = \tuple{\St, \set{\A^{(i)}}, \T, \PS, \gamma, \R}$ takes as given a set of states $\St$ and a set of actions $\A^{(i)}$ for each player. Given a state and actions for each player, the \emph{transition function} $\T : \St \times \A^{(1)} \times \dots \times \A^{(n)} \rightarrow \dist{\St}$ specifies the distribution over next states. The initial state distribution is given by $\PS : \dist{\St}$, while the discount is given by $\gamma$. The shared objective is defined by the \emph{reward function} $\R : \St \times \A^{(1)} \times \dots \times \A^{(n)} \rightarrow \Reals$.

Unlike the case with a regular MDP, in a multiagent MDP the history of interaction can be important, in order for each agent to learn about the other agents. A \emph{history} $h_t : (\St \times \A^{(1)} \times \dots \times \A^{(n)})^t$ encodes the past interaction in the environment. Agent $i$'s \emph{policy} $\pi^{(i)} : H \times \St \rightarrow \dist{\A^{(i)}}$ specifies how the $i$th agent selects actions, and is allowed to depend on history.

Given policies $\set{\pi^{(i)}}$ for all agents, a \emph{trajectory} $\tau$ can be sampled as follows: sample $s_0$ from $\PS$, and then repeatedly sample actions $\set{a^{(i)}_t}$ from $\set{\pi^{(i)}}$ and sample the next state $s_{t+1}$ from $\T$. The shared objective is to maximize the expected reward, which is given by $\expect{\tau}{\sum_t \gamma^t \R(s_t, a^{(1)}_t, \dots a^{(n)}_t)}$.

If we are given every policy except for the $j$th policy $\set{\pi^{(i)} : i \neq j}$, then from agent $j$'s perspective, the other agents can be thought of as ``part of the environment''. In particular, we can reduce the problem of finding the optimal $\pi^{(j)}$ to a single-agent \emph{partially observable} MDP, in which the other policies are a hidden variable of the state. In this POMDP, the transition function $\T'$ samples actions from all the other policies and passes them to $\T$, and the observations include both the original state as well as the actions taken by all of the other agents.

\subsection{Human-AI collaboration via deep RL}  \label{sec:human-ai-collab-deeprl}

In human-AI collaboration, we are given a two-player MDP $\M$, in which the human is one of the players and the AI agent is another. Given just this, it is unclear what the agent should do: the optimal policy for the agent depends heavily on the human's policy, which the agent has no control over. If the human policy $\pi^{(H)}$ is known, then we can simply embed the policy in the environment (as in Section~\ref{sec:multiagent-mdp}) and use reinforcement learning to learn an optimal policy for the agent. Thus, one approach would be to \emph{learn} a human model $\hat{\pi}^{(H)}$, embed the model into the environment, and then use deep RL to find an optimal policy for that model. If $\hat{\pi}^{(H)}$ is sufficiently close to $\pi^{(H)}$, the agent policy will play well with the real human as well \citep{carroll2019utility}.

\section{Unit tests for robustness}

Accurately evaluating robustness is challenging. The typical method of evaluation is to report the average reward on the training distribution, but such an approach does not reveal the low-probability failure modes that are key to evaluating the robustness of an agent to the novel situations it can encounter during deployment.

Even playing our agents with real humans and recording the average reward is not representative of the performance at \emph{deployment}, since in many realistic domains there is a long tail of unusual edge cases that would likely not be seen for any reasonable sample size. In preliminary experiments on Overcooked where we evaluated robustness by playing with real human partners, we found that the noise and differences in human behavior made it very difficult to extract signal out of those tests without huge sample sizes, which motivated us to seek a different evaluation paradigm. Our approach is to place the trained agents in a variety of hand-designed \emph{unit tests}.

\prg{Identifying the relevant inputs.} We first identify the set of inputs in which we should look for edge cases. Human-AI collaboration is specified by a multiplayer MDP $\M = \tuple{\St, \set{\A^{(i)}}, \T, \PS, \gamma, \R}$ and a human model $\pi$; edge cases for the trained agent could be found in $\St$ and in $\pi$. As a result, we develop unit tests that check for edge cases in $\St$ (which we call \emph{state robustness tests}) and for edge cases in $\pi$ (which we call \emph{agent robustness tests}). Intuitively, a test measures \emph{robustness to states} when the success criterion of the test would be approximately independent of the partner model $\pi$. In other words, if the partner model of the test was changed, the success criterion would stay the same (see Figure~\ref{fig:state-robustness} for an example). In contrast, a test measures \emph{robustness to agents} when the success criterion \emph{does} depend on the type of partner. For improved granularity, we subdivide agent robustness tests into those that require history (to identify the type of partner) and those that do not (where the expected behavior is robust to multiple partner types). We name these two categories as ``agent robustness'' (e.g. Figure~\ref{fig:agent-robustness}) and ``agent robustness with memory'' (e.g. Figure~\ref{fig:agent-memory-robustness}).

Given this categorization, our methodology is as follows:
\begin{enumerate}
    \item Identify qualitative situations for each test category;
    \item Concretize each situation to a unit test;
    \item Improve test coverage by observing and probing the trained agents.
\end{enumerate}

\prg{Identifying qualitative situations.} For each of these categories, we brainstorm different possible qualitatively distinct examples of the property under investigation. For example, for state robustness, we can think about states in which soups have been cooked but not delivered, states in which there are many objects on the counters, states in which the agent is holding a useless item, etc. For agent robustness, we might consider cases where the partner is an expert, or where the partner has just started to learn the game, or where the partner has a preference for delivering soups, or where the partner has temporarily stopped playing.

\prg{Concretizing to a unit test.} We then take each of these qualitative situations and aim to distill it down into one or two concrete instances in which it would be obvious to a human how the agent should behave (given enough time to plan). The combination of the concrete situation and the expected correct behavior (within some time limit) then forms one of our unit tests. For example, the test in Figure~\ref{fig:state-robustness} checks whether the agent picks up and places an onion when it is in the best position to do so, while the test in Figure~\ref{fig:agent-robustness} checks whether the agent can cope with a stubborn agent that insists upon delivering the soup it is holding.

Unlike the case in software engineering, where we would normally only have one or two tests for each qualitative setting, we create several instantiations of each qualitative setting with slightly different initial conditions, such as the locations for the counter objects, the agent, and the agent's partner. This is necessary because (unlike typical computer programs) the behavior of a learned agent can vary significantly based on these ``irrelevant'' factors, and so by testing against multiple variations we can significantly reduce the variance of our evaluation.

In some cases, it is challenging to distill the qualitative situation into a concrete one where the correct behavior is unambiguous. For the sake of simplicity, we discard such situations (which we found were relatively rare), since we found that the remaining unit tests were more than sufficient to evaluate current deep RL agents. As agents become more robust, it may become necessary to include such situations in our unit tests as well, potentially by specifying a concrete situation and counting the test as passed if the agent does one of a number of ``reasonable'' behaviors.

\prg{Improving test coverage.} Of course, many environments (including Overcooked) are sufficiently complex that it is nearly impossible to come up with a complete set of edge cases to test agents on. To improve coverage of possible edge cases, it is also useful to look at the behavior of real agents, to inspire new potential edge cases. Concretely, we train a number of deep RL agents (using the different training methods discussed below in Section~\ref{sec:robustness_approaches}). We then watch some games between the agents and directly play several games with them (probing at off-distribution parts of the state space, and behaving in ways that would be uncommon for humans). From this anecdotal data, we extend our list of qualitative settings that a robust agent should successfully navigate, and convert them into unit tests using the methodology above.

\prg{Validating the resulting tests.} As a sanity check, we took the parameterized Theory of Mind agent from Section~\ref{sec:ToM} and set its parameters to make it maximally capable. Since this agent is based on a human-designed planning process, we expect that compared to the deep RL agents it will be much more robust, but potentially worse on average-case performance. We expected it to perform well on the state and agent robustness tests, but not on the memory test, since the agent does not have any state and thus cannot depend on memory\footnote{Technically, the ToM agent does use history for one purpose: when it has been ``stuck'' in a state for some time, it will take random actions to get unstuck. However, this should not be expected to significantly improve performance on memory robustness tests.}. The agent achieved an 86\% score on state robustness tests, and a 95\% score on agent robustness tests, which are quite high in absolute terms and much higher than the deep RL agents (as we will see in the experiments). It also achieved a 75\% score on agent robustness with memory tests, despite not having any state, which is also higher than deep RL agents (which do have state, as they use recurrent neural nets). This suggests that the tests are capturing ``reasonable behavior'' in a wide variety of situations.

We iterated on this methodology for creating unit tests using the layouts from~\citet{carroll2019utility}. This of course runs the risk that our methodology is overfit to these layouts, and so we designed four new layouts, illustrated in Figure~\ref{fig:layouts}, and created a suite of unit tests for these layouts based on the same approach. The full set of unit tests for these layouts is given in Table~\ref{tab:unit_tests}.

Note that even a perfect score on these tests does \emph{not} imply that the tested agent is robust. Since the tests do not exhaustively cover all possible situations, there may still remain failure modes that were not tested for. However, we find that existing agents get fairly low scores on our test suite, suggesting that the test suite can serve as a useful metric for the foreseeable future.

Our unit tests, which are reusable and extensible, are available at \url{https://github.com/HumanCompatibleAI/human_ai_robustness}.

\begin{table*}[t]
  \caption{The suite of unit tests for robustness. The acronym in brackets at the beginning of each \emph{test setup} refers to the test category: SR = State robustness; AR = Agent robustness; A+MR = Agent \& memory robustness. R refers to the agent being tested (the ``robot''), and H refers to the partner agent in the test (the ``human'').}
  \label{tab:unit_tests}
  \begin{tabular}{ll}\toprule
    \textit{Test setup} & \textit{R's response for success} \\ \midrule
    (SR) Soup on counter that should be picked up \& delivered & Pick up and deliver the soup \\
     (SR) Dish or onion on counter, which R requires (Fig.~\ref{fig:state-robustness}) & Pick up said object \\
     (SR) R is holding the wrong object, given the circumstances & Drop object onto the counter \\
     (SR) R is placed in an unlikely location in the circumstances & Adapt \& keep playing as normal \\
     (SR) There's an unlikely number of objects on the counters & Play as normal ( Fig. \ref{fig:state-robustness}) \\ \bottomrule
    (AR) R requires specific object; H blocks relevant dispenser &  Pick up said object from counter \\
    (AR) R holding same object as H, but H is closer to using it & Drop their object onto counter \\
    (AR) R is blocking the path of H & Move out the way (see Fig.~\ref{fig:agent-robustness}) \\ \bottomrule
    (A+MR) H remains stationary (see Fig.~\ref{fig:agent-memory-robustness}) & Adapt and take over H's tasks  \\
    (A+MR) H takes random actions & Adapt and take over H's tasks  \\ \bottomrule
  \end{tabular}
\end{table*}

\section{Robustness through quality and diversity} \label{sec:robustness_approaches}

We trained our deep RL agents using Proximal Policy Optimization (PPO) \cite{schulman2017proximal}; we utilised the open-source implementation by \citet{carroll2019utility}, following their same procedure unless explicitly stated otherwise. One key difference is that we use recurrent neural networks throughout our experiments. For further details of our PPO training, see Appendix~\ref{apdx:PPO}.

The starting point we consider for achieving greater robustness is to partner the deep RL agent, during training, with a human model trained by behavior cloning (BC) on human-human gameplay data (see Section~\ref{sec:BC} below). However, there are a number of problems with such an approach. First, BC only produces a human-like policy on the distribution it was trained on, and can suffer from compounding errors when it moves off-distribution~\cite{ross2011reduction}. In any such area, the learned policy will probably not collaborate well with real humans. Second, not all humans are the same: an agent must be robust to variation across humans if it is to perform well with an arbitrary human; training with a single human model will not incentivize this. These two problems will be addressed in Sections~\ref{sec:ToM} and \ref{sec:pops}.



\subsection{Human model via behavioral cloning} \label{sec:BC}

Behavior cloning learns a policy from expert demonstrations by directly learning a mapping from observations to actions using supervised learning \cite{bain1995framework}. In our case this is a traditional classification task, since we have a discrete action space. Our model takes an encoding of the state as input, and outputs a probability distribution over actions; the model is then trained using the standard cross-entropy loss function.

\begin{figure*}
    \centering
    \includegraphics[height=.18\linewidth]{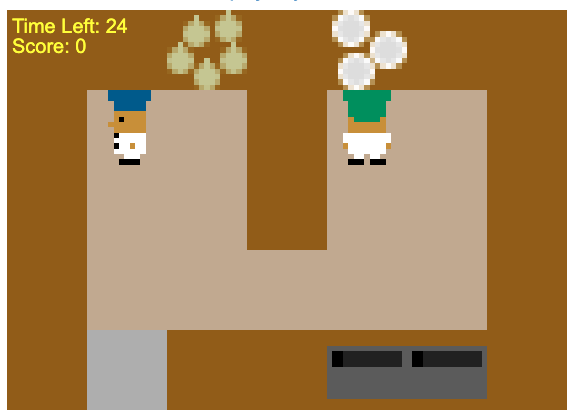}
    \hfill
    \includegraphics[height=.18\linewidth]{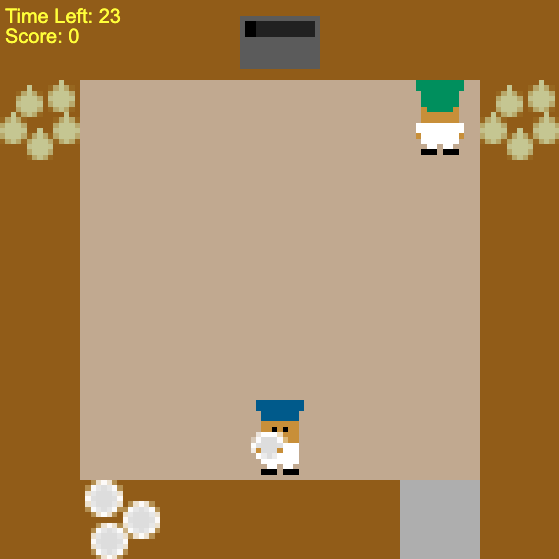}
    \hfill
    \includegraphics[height=.18\linewidth]{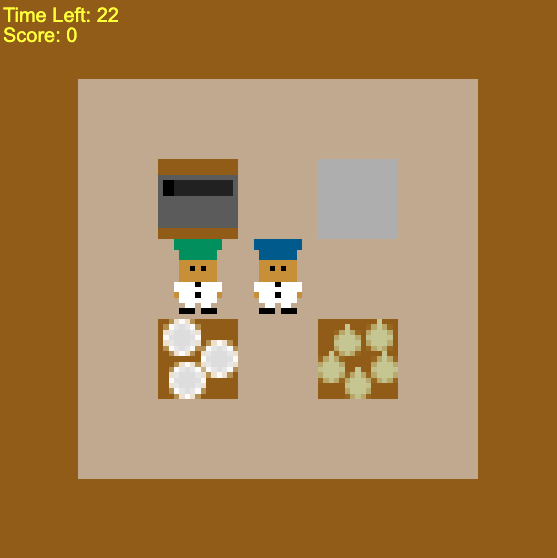}
    \hfill
    \includegraphics[height=.18\linewidth]{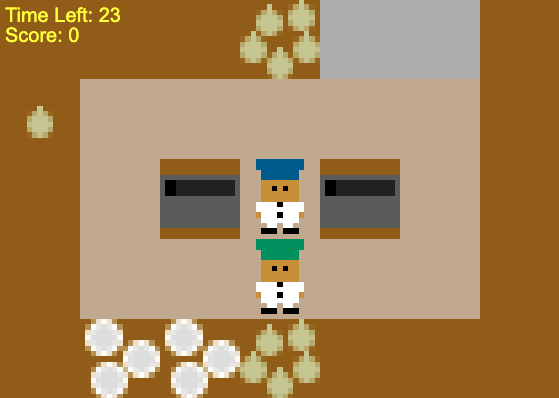}
    \caption{\textbf{Experiment layouts.} From left to right: \emph{Bottleneck} requires frequent travel through a small corridor that only one agent can pass through at a time, leading to challenging motion coordination problems. \emph{Room} is a large empty space, in which gameplay is fairly easy. In \emph{Center Objects}, the most efficient route for moving objects around is through the center, but typically only one agent can use it at a time, thus requiring coordination. \emph{Center Pots} make it easy for agents to interact with pots, and so the primary challenge is in how to navigate to objects around the pots.}
    \label{fig:layouts}
\end{figure*}

For human data collection and behavior cloning training we utilised the open-source implementation by \citet{carroll2019utility}, following their same procedure unless explicitly stated otherwise. As described in Section~\ref{subsection:experimental_setup}, in this paper we used both pre-existing Overcooked layouts from \citet{carroll2019utility}, and created four new layouts (Figure~\ref{fig:layouts}). For the previously existing layouts, we used the pre-existing open-sourced data and behavior cloning models. Instead, for the new layouts of Figure~\ref{fig:layouts} we collected data from Amazon Mechanical Turk. We then removed trajectories that were under-performing based on a set of heuristics to determine engagement of both players in the game. After removal, we had 28 joint human-human trajectories for Bottleneck environment, 31 for Room, 28 for Center Objects, and 31 for Center Pots.

We divide the joint trajectories into two groups randomly under the constraint that the two groups have similar average reward. We then use the two sets of human trajectories to train two sets of behavior cloning models. The first set is used to create BC populations, and the second is used to make up half of the population used to calculate Validation Reward. The hyperparameters used are reported in Table \ref{tbl:bc_hyperparams} in the Appendix. The unit test scores of the BCs themselves are reported in Appendix~\ref{apdx:BC}.

\subsection{Improving the quality of human models: Theory of Mind} \label{sec:ToM}

One idea to solve the first problem introduced above -- that the BC only produces a human-like policy on the distribution it was trained on -- is to improve the quality of our human models across the entire state space. We test one instantiation of this idea: building a parameterized Theory of Mind (ToM) model, i.e. a human model that is structured to reason about other agents' mental states~\cite{choudhury2019utility}. In particular, in this setting we try to make our ToM agent as human-like as possible – we imbue it with biases we expect humans to have (e.g. not be an optimal planner). By design, any ToM agent will behave sensibly on all parts of the state space (as long as its parameters are within reasonable ranges), and so will not suffer greatly from distribution shift.

At every timestep, our Theory of Mind agent enumerates a list of tasks to be completed (such as ``put an onion in the pot'' or ``deliver the soup''), and decides which task it will pursue. It then chooses a low-level action in pursuit of the goal. We call the former the \emph{strategic} choice and the latter the \emph{motion} choice.

At the strategic level, the parameters control how many tasks forward the agent looks ahead; whether the agent takes into account its partner when planning; whether or not to infer its partner's current subtask from its actions; and whether or not to stick to the plan it made on the previous timestep.

At the motion level, the parameters control the probability with which the agent takes a noop action (mimicking the slowness of human players); whether or not the agent takes some time to ``think'' after finishing a subtask; the compliance of the agent during motion planning; whether or not to take the partner into account when planning a path; the probability that the human takes a suboptimal action (via a temperature parameter for a Boltzmann rational model~\cite{ziebart2010modeling}); and the probability of taking a random action. For more information, see Appendix~\ref{apdx:tom_agents}.

\subsection{Diversity of human models: Populations} \label{sec:pops}

In addressing the second problem mentioned above we consider a simple approach to make our deep RL agents robust to variation in humans: train them with a variety of human models. More specifically, by training the agent with a \emph{population} of human models which encompass a diversity of possible strategic behaviors, we can ensure that the resulting agent is able to adapt to any particular human it plays with. On each episode we randomly pick a human model from the population, then our deep RL agent is trained by collaborating with this human model.

However, in collaborative games, it is not immediately clear how to build such a population. It is not just a matter of making the population arbitrarily diverse: a population of random agents is certainly diverse, but is unlikely to lead to an agent that can collaborate well with humans unless the population is extremely large. Nor can we train adversarially, as has been proposed in other contexts~\cite{pinto2017robust,chalaki2019zero, pattanaik2018robust, shen2019robust}, as the adversary would be far too powerful: in many layouts the adversary could simply block the delivery location, ensuring that the agent can never get reward, thus preventing training entirely. This suggests that the agents in the population need to themselves be human models. 


\prg{Population type.} We use populations of BC models, ToM models, or a mixture of the two. 

\prg{Recurrence.} In order to get good results with a population, it should be possible for the agent to learn within an episode which ``type'' of human it is playing with. This is much easier to do when given access to the history of actions that the human has taken in the past. In order to capture this history we use recurrent neural networks for all our deep RL training procedures.

\subsection{Quality of deep RL: leveraging human-human gameplay} \label{sec:diverse-starts}

Since the previous two approaches focused on the human model, they can be thought of as ``fixing the objective'': any method that tries to learn a best response to a human model would benefit from such approaches, not just deep RL. However, in practice we have found that the policies learned by deep RL are themselves not very robust: for example, they may fail when irrelevant dishes are placed on counters\footnote{This can also be seen with the policies trained by \citet{carroll2019utility} at their demo website.}. We hypothesize that this lack of robustness arises because once the trained policy has found a good strategy for getting reward, it is not incentivized to explore other areas of the state space, and so it fails if the test time agent acts differently than expected and brings it to an unseen state.

While this issue is not unique to human-AI collaboration, we do have a potential solution that is not normally available with deep RL: we have access to human-human gameplay data. One would think that if we make the trained policy robust to the \emph{states} in the human-human gameplay data, that could improve its robustness with real humans. We can accomplish this by sampling the initial state of each episode from the human-human data during deep RL training, a technique we call ``diverse starts'' below. One thing to note is that the effectiveness of this procedure is not necessarily obvious: a case could be made that if human-AI gameplay has a sufficiently different distribution of states than human-human gameplay, diverse starts could be increasing robustness to the wrong states.


\section{Experiments}

Our primary goal with the experiments was to evaluate how useful the test suite is in surfacing information about trained agents (\textbf{(H1)} in Section~\ref{sec:hypotheses}).

\subsection{Experimental setup}\label{subsection:experimental_setup}

Our experiments used the four layouts illustrated in Figure~\ref{fig:layouts}; these layouts were designed to capture a range of strategic and coordination challenges. As mentioned above, each unit test has multiple possible initial states. To evaluate the success criterion, we perform 50 evaluation rollouts on the unit tests, and take the average proportion of successes as the score.

We report both the average score on unit tests, as well as the validation reward for each agent, computed as the average reward the agent obtains when partnered with human models from a suite of 20 validation agents (comprised of 10 held out BC and 10 ToM agents). This validation reward is not meant as a measure of the robustness of the agents to novel situations, but rather as a baseline to compare the unit test suite against. The held-out BC agents were trained on a different subset of the human data, and for the ToM agents we chose the parameters by hand to ensure sufficient diversity in the resulting behavior. 

\prg{Manipulated factors.} In our experiments, we manipulate several different factors:
\begin{enumerate}
    \item Whether or not we use diverse starts (Section~\ref{sec:diverse-starts}).
    \item The size of the population we train with. (Note that population size 1 corresponds to the setting of \citet{carroll2019utility} in which there is no population.)
    \item The composition of agents within the population: BC, ToM, or an equal mix of both.
    \item How the ToM models in a population are chosen.
\end{enumerate}

When using multiple BC agents in a population, all agents are trained on the same human-human gameplay data,\footnote{Note that an alternative method would be to train one BC for \emph{each} human in the human-human data, but we found that it was not feasible to get enough gameplay data for each human to make an effective BC model.} but with different seeds. The BC agent chosen for training with a single BC is the best performing out of the 20 BC agents trained. When using multiple ToM agents on the other hand, there are two variants. In the first variant (manual), as with the ToM validation agents, we chose the parameters by hand to ensure sufficient diversity in the resulting behavior (whilst ensuring no overlap with the validation agents -- see appendix for details). In the second variant (random), the parameters are chosen randomly from the set of possible legal values.

\subsection{Hypotheses} \label{sec:hypotheses}
Our first hypothesis (\textbf{H1}) was that the unit tests and the validation reward would supply different information, and will therefore often not be in agreement. In terms of how the different training regimes would impact robustness, our hypotheses -- which we will evaluate using the unit tests -- were: (\textbf{H2}) using diverse starts would increase state robustness, (\textbf{H3}) using ToM would increase state robustness relative to BC (as it would better approximate human behaviour in a larger portion of the state space than BC), (\textbf{H4}) training with populations of human models rather than a single human model would increase robustness, particularly \emph{robustness to agents}, and (\textbf{H5}) training with a mixture of BC and ToM agents would perform better than each individually (for equal-sized populations).

\subsection{Results}

In this section the experimental results we report and analyze are averaged across the four layouts of Figure~\ref{fig:layouts} (see appendix \ref{apdx:results_by_layout} for results broken down by layout).

\begin{figure*}
    \centering
    \includegraphics[width=1\textwidth]{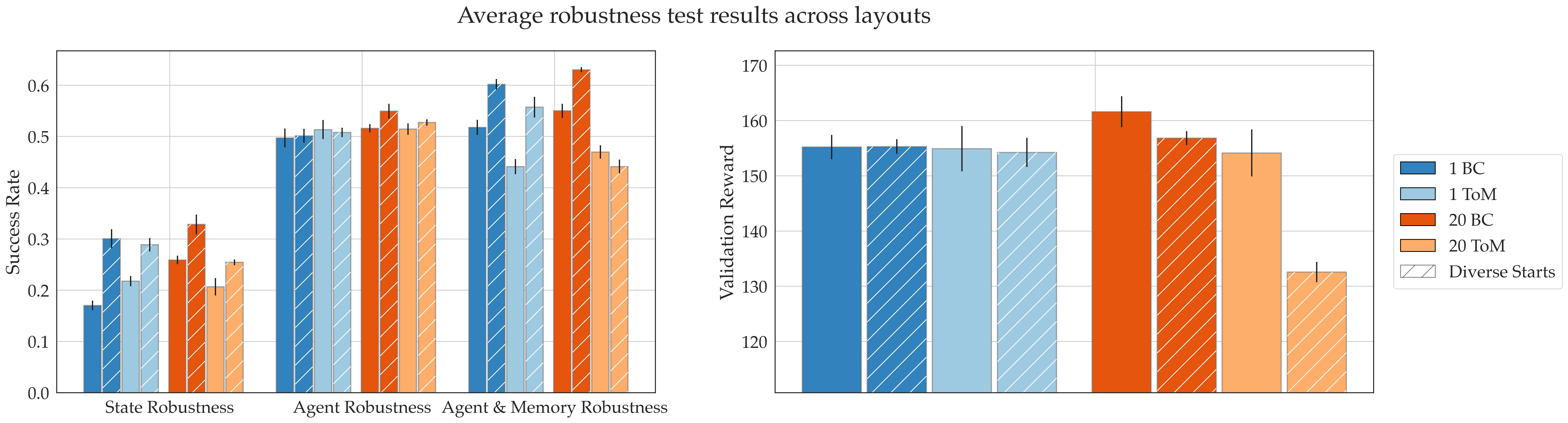}
    \caption{Comparison of robustness scores and validation reward for PPO agents trained with and without diverse starts, with and without a population, and using ToM vs. BC agents. All scores are averaged across the 4 layouts in Figure~\ref{fig:layouts}.}
    \label{fig:all_results}
\end{figure*}

\prg{Diverse starts.} The first thing to notice from Figure~\ref{fig:all_results} is that for diverse starts, the unit tests and the validation reward suggest \emph{opposite} conclusions, in agreement with \textbf{H1} (robustness-reward difference): we see a notable increase in robustness when using diverse starts, for both state and agent robustness tests, but in contrast the validation reward either stays the same (for BC models) or decreases (for ToM models). It appears that using diverse starts confers robustness at the cost of validation reward, an effect that has also been observed with adversarial examples in image classification~\citep{su2018robustness}.

On the axis of robustness, we see that \textbf{H2} (diverse starts) is supported since diverse starts produces an increase in unit test performance across all but one test categories and model types.

\prg{Effect of population vs single.} From Figure~\ref{fig:all_results}, we see that when using BC models, the use of populations leads to improvements across the board for both unit tests and validation reward. However, when using ToM models, we see the opposite effect: the use of populations typically has no effect or hurts performance, again for both unit tests and validation reward. This convergence of unit tests and validation reward is somewhat in opposition to \textbf{H1} (robustness-reward difference), though we note that it is to be expected that some changes will lead to changes in the general ability of the agent and so have similar effects on unit tests and validation reward.

We are not sure what to make of the reversal of the effect for BC and ToM models, and view it as weakly in opposition to \textbf{H4} (populations). One possibility is that since the ToM models are significantly more diverse than the BC models (which only differ based on the initial random seed), a population of ToM models is so diverse that the learning problem becomes too challenging.

\begin{figure*}
    \centering
    \includegraphics[width=1\textwidth]{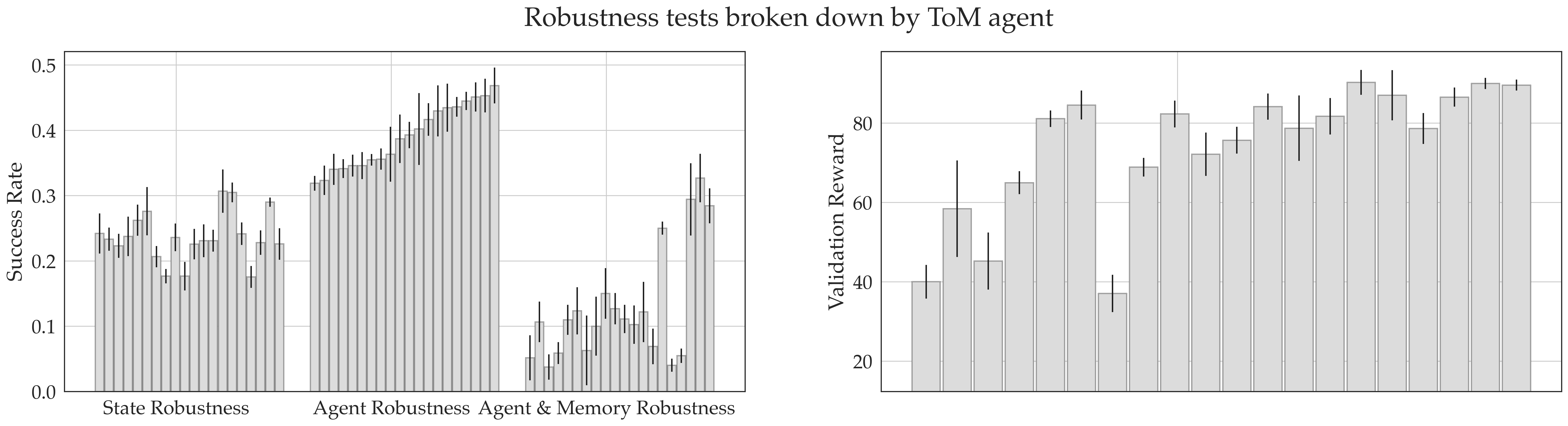}
    \caption{Comparison of robustness scores and validation reward when training with a single ToM agent, for each agent that comprises the ToM population of size 20 (which is used in other experiments). This is performed only on the Counter Circuit layout (see Appendix~\ref{apdx:preliminary_exps} for details). The agents are ordered by increasing agent robustness score. The validation reward and the robustness tests often are substantially different from one another, supporting \prg{H1}.}
    \label{fig:tom_breakdown}
\end{figure*}

To explore this further, we trained an agent with each agent of the manual ToM population individually, reported in Figure \ref{fig:tom_breakdown}, which shows a large variety of robustness and reward across the models, supporting the explanation that too much diversity is a bad thing. It is plausible that this could be fixed with increased model capacity and training time. Nonetheless, we still find the result surprising, and would like to investigate this phenomenon in more detail in future work.

Figure~\ref{fig:tom_breakdown} also shows the importance of selecting a good ToM agent if only training with a single agent (which we can do, by using the MLE estimate for the ToM parameters from the human-human data). We also see that the validation reward and the unit tests can disagree significantly, supporting \textbf{H1} (robustness-reward difference). 

\begin{figure*}
    \centering
    \includegraphics[width=1\textwidth]{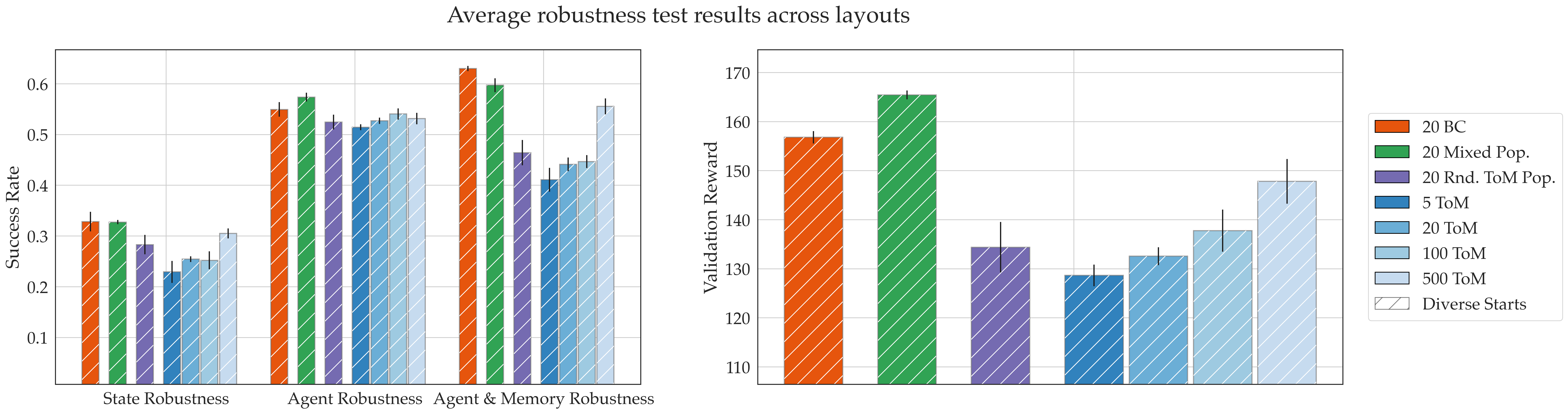}
    \caption{Comparison of robustness scores and validation reward for PPO agents trained with populations of 20 BCs, a mixed population of 10 BCs and 10 ToM agents, a population of 20 random ToM agents, and populations of various sizes of manually selected ToMs. All experiments were performed with diverse starts, and all scores are averaged across the 4 layouts in Figure~\ref{fig:layouts}.}
    \label{fig:population_ablation}
\end{figure*}

\prg{Single BC vs single ToM.} From Figure~\ref{fig:all_results}, we see that when comparing agents trained with a single BC and a single ToM, using the ToM agent can lead to an improvement in state robustness (supporting \textbf{H3} (ToM)), but using diverse starts effectively nullifies this improvement. We speculate that this is because the additional robustness conferred by training with the ToM agent is a subset of that conferred by diverse starts, and so once diverse starts is used ToM no longer provides any benefit.

On all other metrics, including validation reward, we see that a single ToM performs comparably to or worse than the corresponding BC agent. This suggests that by using a ToM agent, we are increasing the number of states on which the agent can perform reasonably, but decreasing the agent's ``average'' capability. We would not be able to make such inferences from just the validation reward, where BC and ToM are nearly identical, again supporting \textbf{H1} (robustness-reward difference).

Looking across all of the settings (instead of just the single BC and single ToM setting), we find that the agents trained with ToM partners tend to perform worse across the board relative to agents trained with BC partners. As before, we speculate that this is because the ToMs add too much diversity to the training, such that the learning problem becomes too challenging.

\prg{Mixed population.} Figure~\ref{fig:population_ablation} illustrates the robustness and reward across different methods of constructing the population of agents. We see that according to the unit tests, using a mixed BC and ToM population is approximately on par with the population of BC agents (contradicting \textbf{H5} (mixed-population)). However, the mixed population leads to a significant increase in validation reward, most likely because the validation population is itself a mixture of BC and ToM agents. Once again we see that our two evaluation metrics provide different insight into the method, in accordance with \textbf{H1} (robustness-reward difference).

The unit tests also suggest that the mixed population has better agent robustness but worse memory robustness than the population of BCs. However, there is no clear reason why this would be the case, and the differences are fairly small. 

\prg{Random ToM population and Effect of population size.} Interestingly, the randomly chosen ToM population actually outperforms choosing parameters manually to maximise ToM diversity. However, any population of ToM agents is significantly outperformed by using a BC or mixed population. Figure~\ref{fig:population_ablation} shows that all metrics increase (albeit relatively slowly) by increasing the population size, up to the maximum size we tested (500).

\section{Limitations and future work}

\prg{Summary.} In this work, we propose the use of testing to evaluate the robustness of collaborative agents. These unit tests search for potential edge cases in possible partner behavior and possible states encountered, in order to reveal when the trained agent will be robust to novel situations during deployment. Using these unit tests, we evaluated three natural proposals for improving robustness in human-AI collaboration via deep RL: improving human model quality through a parameterized Theory of Mind (ToM) agent, training with a diverse population of collaborative agents, and initializing from states visited in human-human gameplay. The unit tests revealed information about the method that was relatively uncorrelated with the average reward metric: sometimes unit test robustness increased at the cost of average reward (as with initialization from states in human-human gameplay), sometimes different types of robustness were affected while average reward stayed the same (as with the use of a single ToM agent as the partner), sometimes unit test robustness remained the same while average reward was improved (as with the use of a mixture of BC and ToM agents), and sometimes unit test robustness and average reward were in sync (as with the effects of using a population). While our best results used a population of 10 BC and 10 ToM agents, we emphasize that our primary finding is that our unit test suite provides information that may not be available by simply considering validation reward, and our conclusions for specific techniques are more preliminary.

\prg{Future work on \emph{evaluating} robustness.} One challenge we encountered was how to evaluate robustness in cases where the correct behavior is ambiguous. As an example in Overcooked, what would be the correct behaviour for the blue player in Figure~\ref{fig:state-robustness}, if the green player was holding nothing (instead of holding a dish)? Fetching either a dish or an onion can be correct, depending on the gamaplay style and preferences of the other player. Future work could address how to create reliable unit tests to measure robustness in these ambiguous scenarios: for example, perhaps a few ``reasonable'' behaviors could be identified, and we could check whether the agent executes one of these ``reasonable'' behaviors. Beyond this, a natural extension of our work is to expand the use of unit tests to other domains besides human-AI collaboration.

\prg{Future work on \emph{improving} robustness.} As mentioned above, we would like to see more work further evaluating our proposals, especially in the case of populations where we found a positive effect for BC but a negative one for ToM. There are also several additional avenues for improving robustness. While in this work we improved the quality and diversity of the partner agent, all of our agents were still trained for a specific layout. Arguably, for true robustness (especially robustness to states), we need diversity in \emph{layouts} as well. An alternative direction for future work is to explore meta learning, in order to train the agent to \emph{adapt} online to the specific human partner it is playing with. This could lead to significant gains, especially on agent robustness with memory.

\begin{ack}
This work was partially supported by Open Philanthropy, Microsoft and NSF CAREER. PK acknowledges support from the Royal Commission for the Exhibition of 1851, and the Foundational Questions Institute (FQXi) under the Intelligence in the Physical World Programme (Grant No. FQXiRFP-IPW-1907). We thank researchers at the Center for Human-Compatible AI and the InterACT lab for helpful discussion and feedback.
\end{ack}

\bibliographystyle{plainnat}
\bibliography{references}

\newpage

\appendix

\section{Experimental results split by layout} \label{apdx:results_by_layout}

\begin{figure*}[h]
    \centering
    \includegraphics[width=1\textwidth]{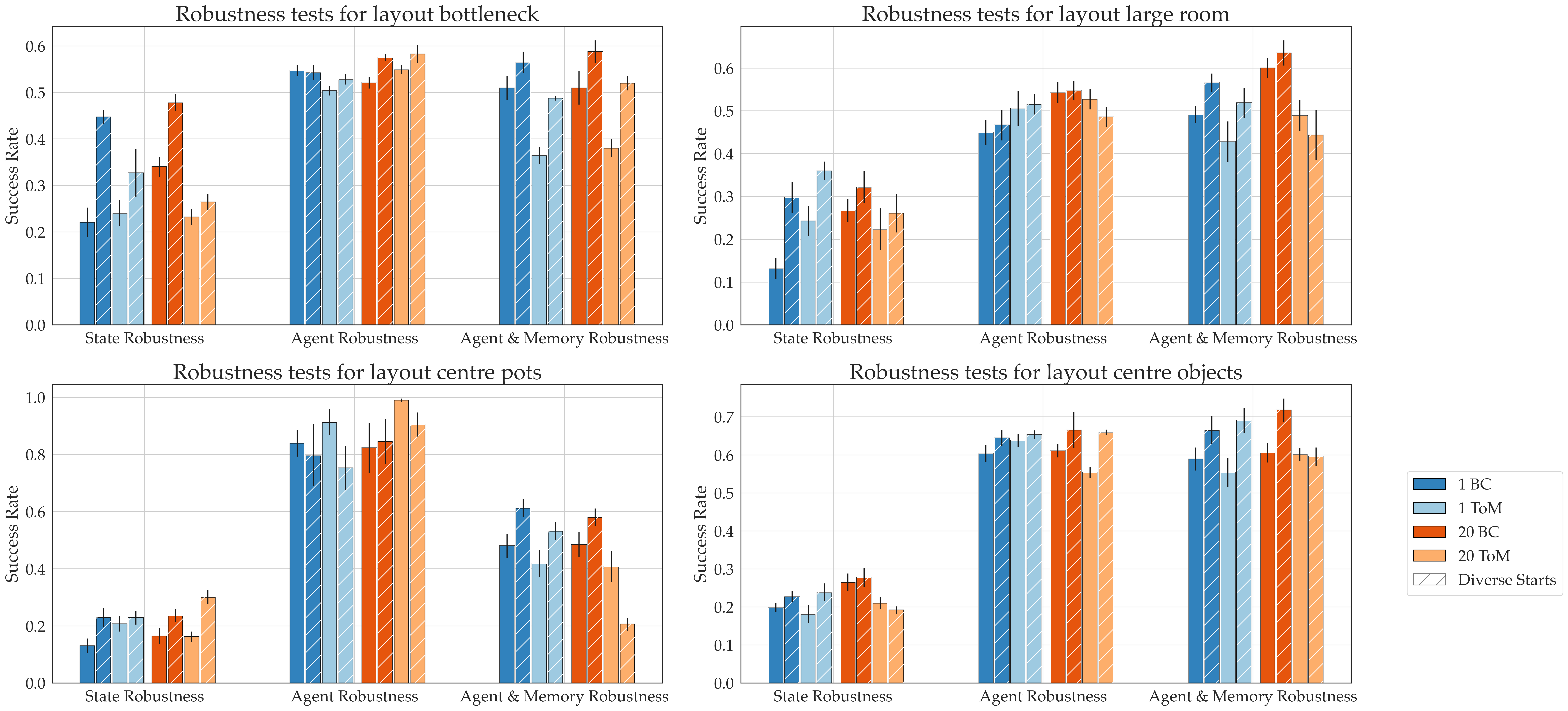}
    \caption{Figure \ref{fig:all_results} (left sub-graph) broken down by layout.}
    \label{fig:all_layout_results_tests}
\end{figure*}

\begin{figure*}[h]
    \centering
    \includegraphics[width=1\textwidth]{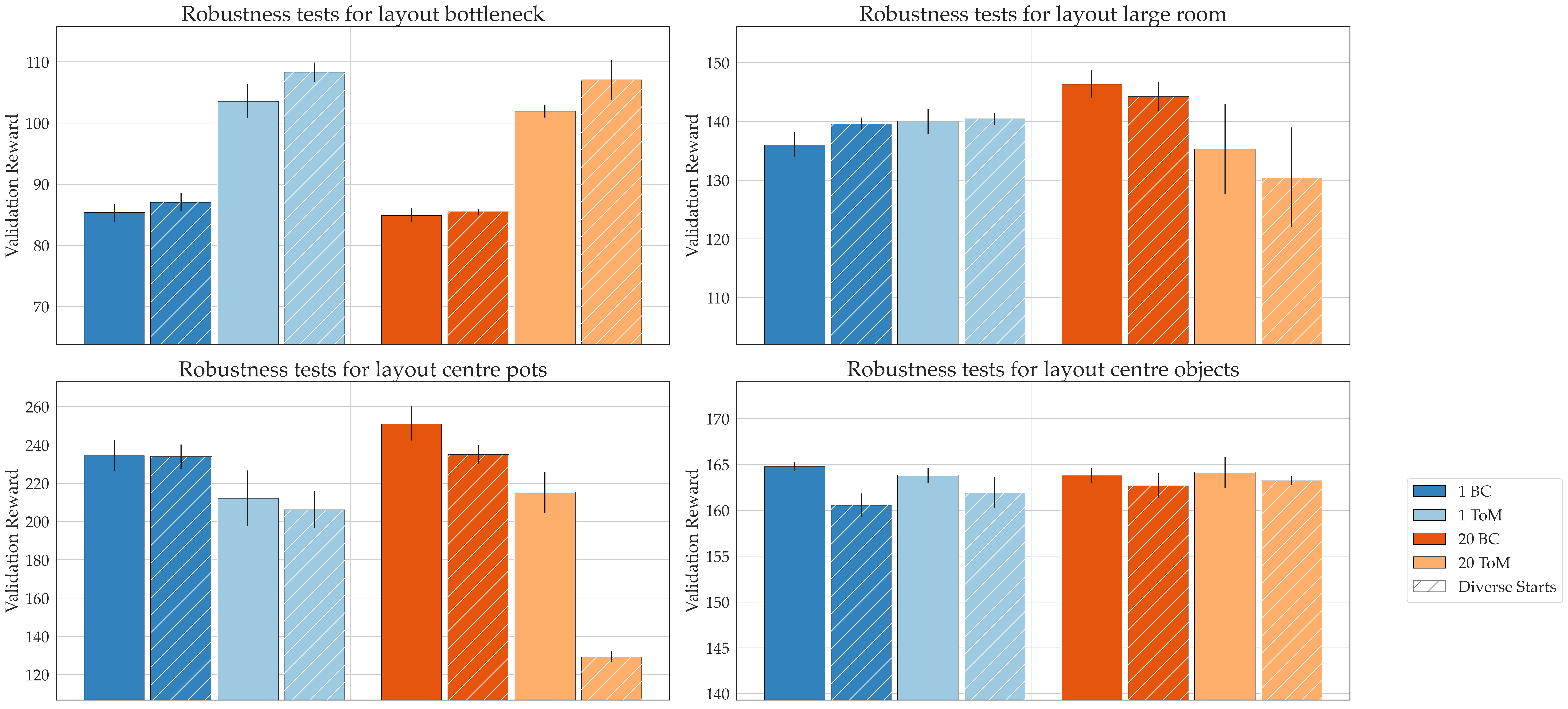}
    \caption{Figure \ref{fig:all_results} (right sub-graph) broken down by layout.}
    \label{fig:all_layout_results_rewards}
\end{figure*}

\begin{figure*}[h]
    \centering
    \includegraphics[width=1\textwidth]{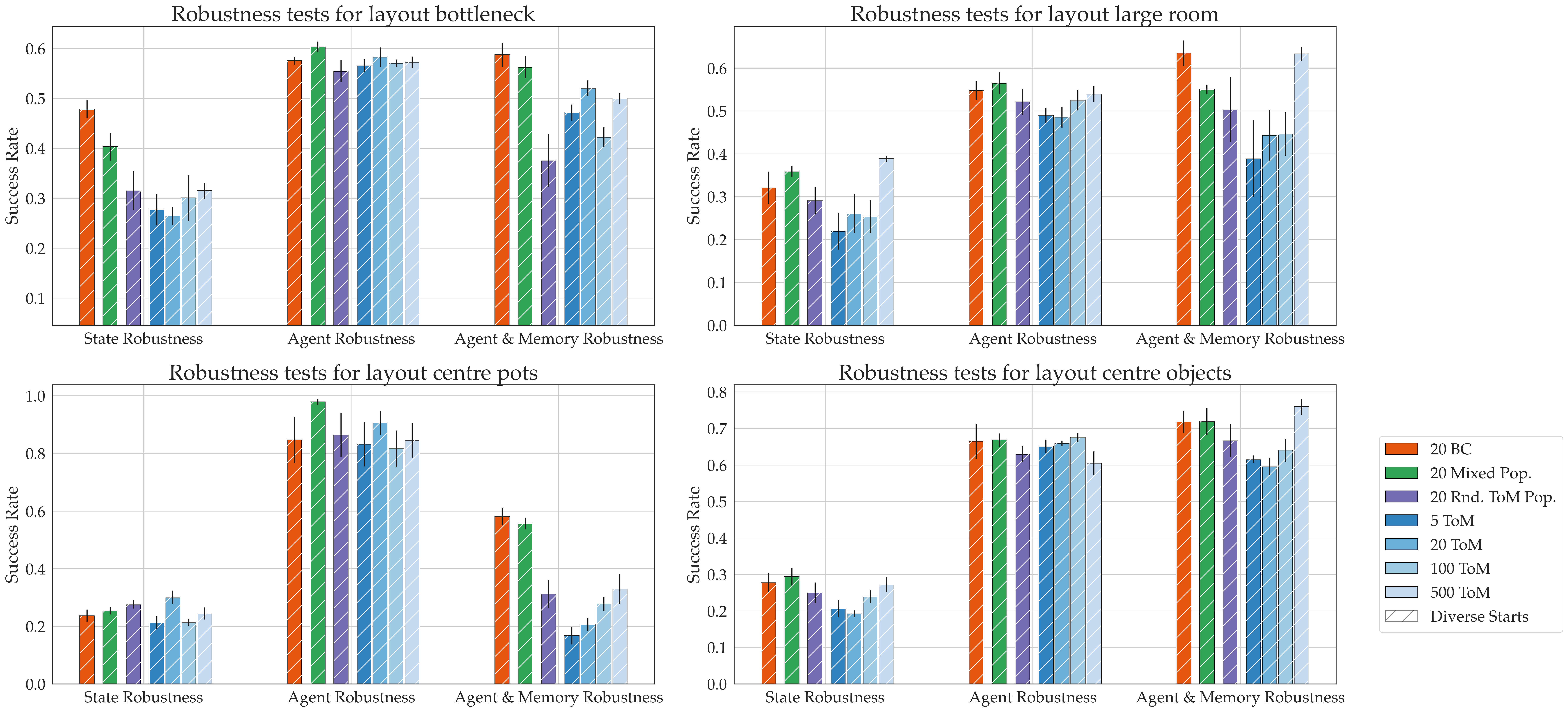}
    \caption{Figure \ref{fig:population_ablation} (left sub-graph) broken down by layout.}
    \label{fig:all_layout_results2_tests}
\end{figure*}

\begin{figure*}[h]
    \centering
    \includegraphics[width=1\textwidth]{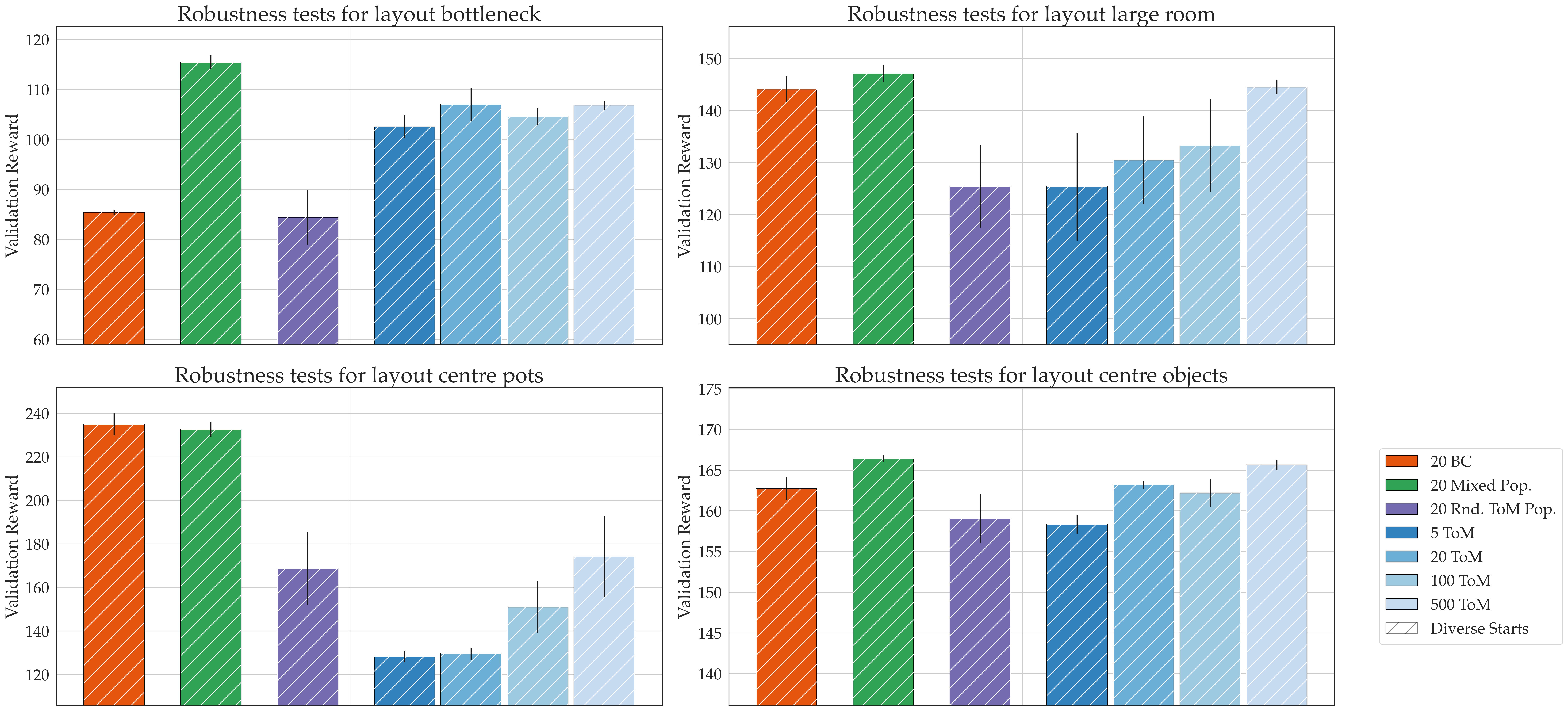}
    \caption{Figure \ref{fig:population_ablation} (right sub-graph) broken down by layout.}
    \label{fig:all_layout_results2_rewards}
\end{figure*}

In Figures~\ref{fig:all_results} and \ref{fig:population_ablation} we only reported the experimental results averaged across the four layouts of Figure~\ref{fig:layouts}. In Figures~\ref{fig:all_layout_results_tests}-\ref{fig:all_layout_results2_rewards} we report our experimental results broken down by layout.


We will now highlight some notable exceptions to the conclusions drawn from looking only at the results that were averaged across layouts. When averaged over layouts, in Figure~\ref{fig:all_results} we saw an increase in robustness when using diverse starts. One notable exception to this is when playing with the ToM agent on Center Pots for the agent robustness tests: here the diverse starts \textit{decreased} the robustness. For the state robustness tests, when averaged over layouts there was no significant difference between 1 BC and 1 ToM (when both had diverse starts). When this is broken down by layouts in Figures~\ref{fig:all_layout_results_tests} and \ref{fig:all_layout_results_rewards}, we see that in Center Objects and Large Room, playing with the ToM was a little more robust than with the BC; whereas for Bottleneck, playing with the ToM was \emph{less} robust. Finally, if we focus just on using ToM populations, we see that, with diverse starts, some noticeable outliers occur for Center Pots. Here, for both the agent robustness with memory tests and the validation reward, the population of 20 ToMs performs significantly worse than the alternatives.


Moving on to Figure~\ref{fig:population_ablation}, we saw that the mixed population receives significantly more validation reward than the BC population. However, we see from Figure~\ref{fig:all_layout_results2_rewards} that this reward improvement is solely down to the strong performance of the mixed population on Bottleneck. Instead comparing the randomly chosen ToM population with choosing the parameters manually, we saw in Figure~\ref{fig:population_ablation} that randomly-chosen outperforms manually-chosen. There are several minor exceptions to this in Figures~\ref{fig:all_layout_results2_tests} and \ref{fig:all_layout_results2_rewards}, and two strong exceptions, in which manually-chosen does significantly better. The latter both occur on Bottleneck, for both the validation reward and the agent robustness with memory tests. Our final comparison in the main text was between the different sizes of ToM populations. In Figure~\ref{fig:population_ablation} we saw that all metrics increased gradually when we increase the population size. Broken down by layout in Figures~\ref{fig:all_layout_results2_tests} and \ref{fig:all_layout_results2_rewards}, again there are several minor exceptions to this, in particular on Center Pots.

Future work will analyse these anomalies seen in Figures~\ref{fig:all_layout_results_tests}-\ref{fig:all_layout_results2_rewards}, and their implications.

\section{Preliminary experimental results for original Overcooked layouts} \label{apdx:preliminary_exps}

\begin{figure*}[h]
    \centering
    \includegraphics[width=1\textwidth]{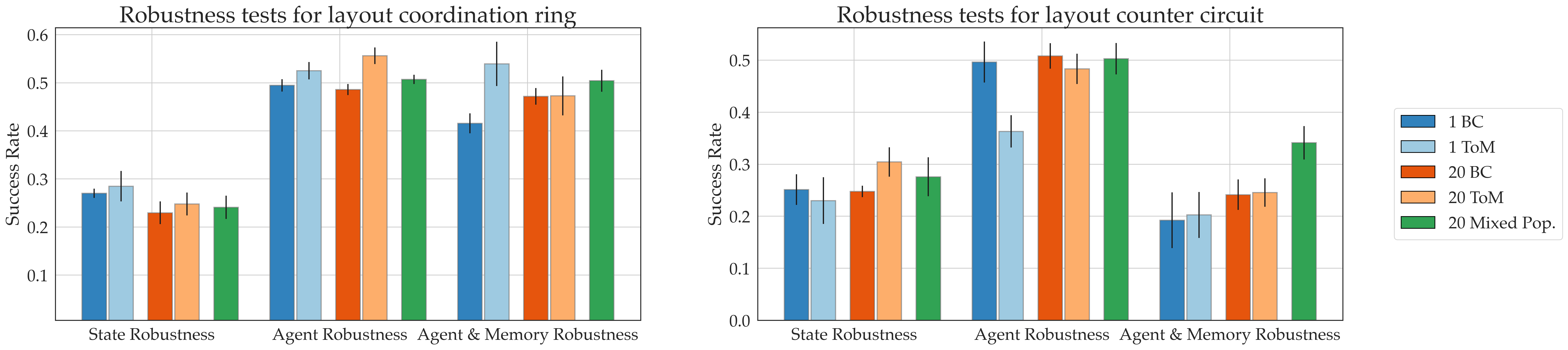}
    \caption{Robustness tests evaluation for two of the original Overcooked layouts.}
    \label{fig:old_layout_results_tests}
\end{figure*}

\begin{figure*}[h!]
    \centering
    \includegraphics[width=1\textwidth]{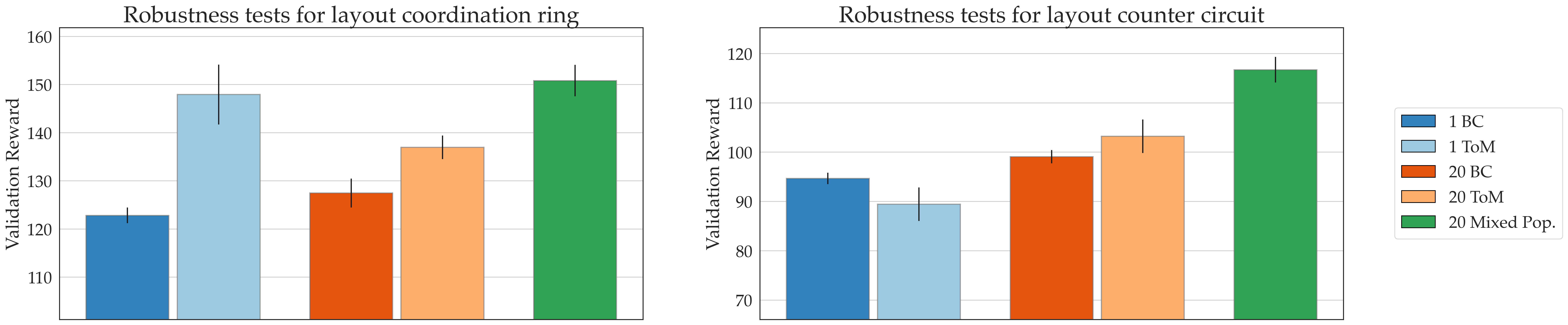}
    \caption{Validation Reward evaluation for two of the original Overcooked layouts.}
    \label{fig:old_layout_results_rewards}
\end{figure*}

Before running the whole suite of experiments described in Section \ref{subsection:experimental_setup} for the 4 layouts of Figure~\ref{fig:layouts}, we did a preliminary experimental investigation on two of the original layouts from \citet{carroll2019utility}. The results of the experiments we conducted – which are just a subset of the final ones – are in Figures~\ref{fig:old_layout_results_tests} and \ref{fig:old_layout_results_rewards}.


\section{Behavior cloning unit test scores} \label{apdx:BC}

\begin{table*}[h!]
\centering
\begin{tabular}{ |p{3.2cm}||p{1.6cm}|p{1.6cm}|p{1.6cm}|p{1.6cm}|p{1.6cm}|  }
 \hline
 \multicolumn{5}{|c|}{Behavior cloning hyperparameters} \\
 \hline
 Parameter & Bottleneck & Room & Center Obj. & Center Pots\\
 \hline
 Learning Rate  &   1e-3    & 1e-3 &   1e-3 & 8e-4\\
 \# Epochs &   130  &     90   & 80 & 180\\
 Adam $\epsilon$  &   1e-8      & 1e-8 &  1e-8 & 1e-8\\
 \hline
\end{tabular}
\vskip 0.1in
\caption{Hyperparameters for behavior cloning across the 4 new layouts.}
\label{tbl:bc_hyperparams}
\end{table*}

We evaluated the best performing BC for each layout on our suite of unit tests (the `best performing' was found by partnering each BC with every other BC, then finding the average reward over partners). Averaged over layouts, these BCs achieved the following success rates: \textit{state robustness tests}: $0.21$; \textit{agent robustness tests}: $0.45$; and \textit{agent robustness with memory tests}: $0.29$. Note that these BCs performed far worse than the ToM agents (the ToM scores are reported in Appendix~\ref{apdx:tom_agents}). The average validation score across layouts was $71.8$.

\section{ToM agents}
\label{apdx:tom_agents}

As introduced in the main text, at every timestep our Theory of Mind agent enumerates a list of tasks to be completed (such as “put an onion in the pot” or “deliver the soup”), and decides which task it will pursue. It then chooses a low-level action in pursuit of the goal.  We call the former the strategic choice and the latter the motion choice.

\noindent\textbf{Strategic choice}

\noindent\emph{Probability of being greedy}: If this is set to 1, then the agent will always do the highest-priority task on the list (or the lowest cost task if there are two equal-priority tasks). In the other extreme, if this is 0 then the agent will always jointly plan the best team strategy in order to complete the first N tasks on the list (the value of N is a separate parameter, “lookahead horizon”, discussed below). For example, if N=2 then the agent will determine the lowest-cost strategy for both agents executing all tasks on the list up to priority 2.

\noindent\emph{Lookahead horizon}: This is the value of N discussed above: i.e. how far to look ahead down the task list when jointly planning.

\noindent\emph{Probability of factoring in the other agent’s current perceived actions}: We determine the other agent’s current perceived action by asking 1) are they already carrying an object that can be used, and if not then 2) is there a (useful) object in their 180 degree field of view that they could pick up. If either of these is true, then we assume that the other agent will complete the associated task, then we cross this task off the task list. We then plan either greedily or jointly, using the revised task list, according to the value of the \emph{probability of being greedy} parameter.

\noindent\emph{Retain-goals}: At each timestep the agent will keep its current goal with this probability. Here “goal” refers to the sub-task level, i.e. dropping an object; picking up an object; using an object. Each time a sub-task is completed then the agent always re-plans. If retain-goals is zero then at every timestep the agent re-plans.

\begin{table*}[h]
\centering
\begin{tabular}{ |p{2.7cm}||p{1.6cm}|p{1.6cm}|p{1.6cm}|p{1.6cm}|  }
 \hline
 \multicolumn{5}{|c|}{PPO hyperparameters} \\
 \hline
 Parameter & Bottleneck & Room & Center Obj. & Center Pots\\
 \hline
 Learning rate  &   5e-4    & 5e-4 &   1e-3 & 1e-3\\
 SP anneal. horizon  &  [3e6,1e7]    &  [3e6,7e6] & [3e6,1e7] & [3e6,1.3e7] \\
 \hline
\end{tabular}
\vskip 0.1in
\caption{Hyperparameters for PPO across the 4 new layouts.}
\label{tbl:ppo_hyperparams}
\end{table*}

\noindent\textbf{Motion choice}

\noindent\emph{Prob-pausing}: This is the probability of pausing in a given timestep. This is needed because human players often “pause” quite a lot simply because they can’t press the keys / think fast enough to make a move on every single timestep.

\noindent\emph{Thinking-prob}: After achieving a sub-task (e.g. picking up an onion) the agent waits to "think". This simulates the fact that humans will often spend more time pausing to think after they have achieved a sub-task.

\noindent\emph{Compliance}: In games with limited space to move, players will often bump into each other. When this happens, compliance is the probability that the agent will take an avoiding action (e.g. step backwards).

\noindent\emph{Path-teamwork}: Once the agent has a goal, it can choose different paths to reach the goal. Here, path-teamwork is the probability of factoring in the other player when finding the best path (i.e. they use a joint motion planner).

\noindent\emph{Rationality coefficient}: Humans will not always take an action along the shortest path to a goal. The ToM takes a sub-optimal action with a Boltzmann rational probability. Setting to 0 means they always take random actions; setting to $\infty$ means they always take the lowest cost path (in practice setting to 20 is large enough for the latter).

\noindent\emph{Prob random action}: At each timestep the agent will take one of the 6 random actions with this probability.

\textbf{Choosing the ToM parameters manually:} Our intention here was to make ToM agents that behave like human players, whilst ensuring as much diversity as possible when creating populations of ToM agents. We first made a population of ToMs by fitting all parameters to human-human data, using a metropolis sampling algorithm. For population size 1, we used the maximum likelihood set of parameters for the ToM agent. However, for larger populations, this sampling procedure resulted in a quite uniform population. So, to increase the diversity, we instead decided to manually selected ToM parameters for the population, building from the fitted parameters. Throughout this process, we played several games with different ToM agents to ensure they still had human-like gameplay and skill levels. The values of \emph{prob-pausing} were selected in the range of $\pm0.2$ from the values from the metropolis sampling (which varied across layout, but were 0.6 on average across layouts). \emph{Lookahead horizon} ranged between 1 and 4; \emph{Rationality coefficient} ranged from 1 and 20; \emph{Thinking-prob} ranged from 0 to 0.4; and \emph{Prob random action} was set to 0. For all other parameters, the full allowed range of $[0,1]$ was used, and as all such values lead to human-like behavior.

\textbf{ToM agents' unit test scores:} We evaluated the maximum likelihood ToM agent for each layout on our suite of unit tests. Averaged over layouts, these ToM agents achieved the following success rates: \textit{state robustness tests}: $0.81$; \textit{agent robustness tests}: $0.74$; and \textit{agent robustness with memory tests}: $0.53$. The average validation score across layouts was $77.7$.

\section{PPO agent training} \label{apdx:PPO}

For PPO we built off of the open-source implementation by \citet{carroll2019utility}, following their same procedure unless explicitly stated otherwise. One difference we would like to highlight is that we use recurrent neural networks throughout our experiments. We started with their hyperparameter choices as initial values, then tuned each parameter, focusing largely on total batch size, learning rate, reward shaping horizon (as in \citep{carroll2019utility}, we augment the reward with a denser reward signal to facilitate convergence), and SP annealing horizon, varying only 1 or 2 hyperparameters together each time, until we converged on the hyperparameters reported below.

We parameterize the policy with a convolutional neural network with 3 convolutional layers, each of which has 25 filters, followed by 3 fully-connected layers with hidden size 64. Hyperparameters common to all layouts are: entropy coefficient (= 0.1), gamma (= 0.99), lambda (= 0.98), clipping (= 0.05), maximum gradient norm (= 0.1), gradient steps per minibatch per PPO step (= 8), \# minibatches (= 15), learning rate annealing (= 1.5), reward shaping horizon (= 2e7), total batch size (= 48000). Layout-specific hyperparameters are reported in Table \ref{tbl:ppo_hyperparams}. We ran the PPO training until the training reward failed to improve for 5e6 timesteps.

All PPO agents are trained on 5 seeds. Each seed is then evaluated with our suite of unit tests and with the validation population. All graphs report results with mean and standard errors across seeds.

Our PPO agents were trained, 8 runs in parallel, on an Azure cloud instance with 4x NVIDIA Tesla K80 cards and 24x Intel Xeon E5-2690 v3 processors. Each run took 16-48 hours, with the large variability due to the fact that we trained until convergence of the reward.

\end{document}